\pdfoutput=1
\PassOptionsToPackage{table}{xcolor}
\documentclass[11pt]{article}

\usepackage[final]{acl}
\usepackage{afterpage}
\usepackage{times}
\usepackage{latexsym}
\usepackage{listings}
\usepackage{graphicx}
\usepackage{subcaption}
\usepackage[T1]{fontenc}
\usepackage[utf8]{inputenc}
\usepackage{microtype}
\usepackage{inconsolata}
\usepackage{graphicx}
\usepackage{booktabs}  

\usepackage{latexsym}
\usepackage{adjustbox}
\usepackage{xspace}
\usepackage[table]{xcolor}
\usepackage{todonotes}
\usepackage{comment}
\usepackage{tikz}
\usepackage{pgfplots}
\usepackage{pgfplotstable}
\usepackage{makecell}
\usepackage{adjustbox}
\usepackage{pifont}
\usepackage{float}
\usepackage{supertabular}
\usepackage{multirow}
\usepackage{longtable}
\usepackage{graphicx}
\usepackage[edges]{forest}
\definecolor{hidden-draw}{RGB}{0,0,0}
\usepackage{wrapfig}
\usepackage{CJKutf8}
\usepackage{CJK}

\usepackage{tcolorbox}
\tcbset{
    mybox/.style={
        fontupper=\linespread{.8}\selectfont,
        sharp corners,
        top=0pt, 
        bottom=0pt 
    }
}

\usepackage{amsmath}
\usepackage{amssymb}

\usepackage{booktabs}
\usepackage{threeparttable}
\usepackage{tabularx}

\usepackage{soul}

\usepackage{bm} 

\usepackage{algorithm}
\usepackage{algorithmic}

\usepackage{setspace}

\newcommand{\avgstd}[2]{#1{\scriptsize\ensuremath{\pm}#2}}
\newcommand{\errbest}[1]{\cellcolor{celadon!8}#1}
\newcommand{\errlow}[1]{\cellcolor{celadon!16}#1}
\newcommand{\errmid}[1]{\cellcolor{celadon!24}#1}
\newcommand{\errhigh}[1]{\cellcolor{celadon!32}#1}

\definecolor{mygreen}{RGB}{11,141,10}
\definecolor{myred}{RGB}{223,68,52}
\definecolor{myblue}{RGB}{70,130,180}
\definecolor{mydeepblue}{RGB}{65,105,225}
\definecolor{myviolet}{RGB}{97,0,138}
\definecolor{myburgundy}{RGB}{110,10,30}
\definecolor{myblue2}{RGB}{0,105,148}
\definecolor{iceblue}{RGB}{173, 216, 230}
\definecolor{puregreen}{RGB}{0, 218, 0}
\definecolor{graygreen}{RGB}{74,113,106}

\definecolor{wingreen}{rgb}{0,0.45,0.24}
\definecolor{losered}{rgb}{1.0,0.1,0.24}

\definecolor{lightcoral}{rgb}{0.97, 0.36, 0.46}
\definecolor{lightyellow}{rgb}{0.98, 0.7, 0}
\definecolor{harvestgold}{rgb}{0.85, 0.57, 0.0}
\definecolor{brightlavender}{rgb}{0.75, 0.58, 0.89}
\definecolor{capri}{rgb}{0.0, 0.75, 1.0}
\definecolor{carminepink}{rgb}{0.92, 0.3, 0.26}
\definecolor{celadon}{rgb}{0.67, 0.88, 0.69}
\definecolor{darkpastelgreen}{rgb}{0.01, 0.75, 0.24}

\definecolor{grayhighlight}{RGB}{250,250,227}

\definecolor{target}{HTML}{F47983}
\definecolor{control}{HTML}{3E87CD}
\definecolor{credibility}{HTML}{B98AC9}
\definecolor{logical}{HTML}{93C572}
\definecolor{emotional}{HTML}{F9EAC3}

\newenvironment{packeditemize}{
\begin{list}{$\bullet$}{
\setlength{\labelwidth}{8pt}
\setlength{\itemsep}{0pt}
\setlength{\leftmargin}{\labelwidth}
\addtolength{\leftmargin}{\labelsep}
\setlength{\parindent}{0pt}
\setlength{\listparindent}{\parindent}
\setlength{\parsep}{0pt}
\setlength{\topsep}{3pt}}}{\end{list}}

\usepackage{pifont}

\newcommand{\myline}{\par
  \kern0pt 
  \hrule height 0.6pt
  \kern3pt 
}

\newcommand{\mylinenoskip}{\par
  \kern3pt 
  \hrule height 0.6pt
  \kern3pt 
}

\usepackage{amsthm}

\usepackage{url}
\usepackage{booktabs}
\usepackage{siunitx}
\title{Can Released LLM Vocabularies Support Token-Level Estimation of Hidden Corpora?}

\author{
    \centerline{Qingjie Zhang\textsuperscript{1,2}, \ 
    Xingzhang Ren\textsuperscript{2*}, \ 
    Zixuan Chen\textsuperscript{1}, \ 
    Jinfeng Li\textsuperscript{3}, \
    Yuefeng Chen\textsuperscript{3},
    } \vspace{0.5mm}  \\
    \centerline{\textbf{
    Yitong Yang\textsuperscript{3}, \ 
    Hui Xue\textsuperscript{3}, \
    Dayiheng Liu\textsuperscript{2*}, \ 
    and Han Qiu\textsuperscript{1*}
    }} \vspace{0.5mm} \\
    \centerline{\normalsize{$^{1}$Tsinghua University~~$^{2}$Qwen Team, Alibaba Group~~$^{3}$Alibaba Group}} \vspace{0.5mm} \\
    \centerline{\normalsize{Emails: \{qj-zhang24@mails., qiuhan@\}tsinghua.edu.cn~~\textsuperscript{*}Corresponding authors}}
}

\begin{document}
\begin{CJK}{UTF8}{gbsn} 
\maketitle
\begin{abstract}
Pretraining corpus composition shapes LLM capabilities, but it often remains hidden even when model weights are released. Prior work has inferred corpus mixtures or traced specific token groups from released tokenizer vocabularies; in contrast, we estimate corpus ratios for arbitrary target tokens. We first show that BPE tokenizers trained on different corpora share stable token ID--ratio distributions, motivating distribution transfer from known corpora to a target tokenizer trained on hidden corpora. We then propose Quantile-Guided Density Estimation (QGDE), which approximates this distribution with multiple quantile trends and uses local density weighting to produce token-level estimates. In controlled settings and a realistic setting using the released SmolLM tokenizer, QGDE achieves mean relative errors as low as 3.00\% for token-level estimation and 3.08\% after aggregation into category-level mixtures. These results suggest that released tokenizer vocabularies provide a useful signal for fine-grained corpus estimation beyond coarse composition inference. The code of QGDE is available at \url{https://github.com/qingjiesjtu/QGDE}.
\end{abstract}

\section{Introduction}
\label{sec:intro}

The composition of pretraining corpora is essential for interpreting LLM performance, since it shapes basic capabilities such as multilingual and domain-specific performance \citep{xie2023doremi,hoffmann2022training,li2024datacomp,petty2025codepretraining}. However, even models with released weights often leave the corpus opaque \citep{touvron2023llama2,deepseek2024deepseek,liu2024deepseek}. A more accessible signal is the tokenizer vocabulary, which is often released and commonly trained with byte-pair encoding (BPE) \citep{sennrich-etal-2016-neural} on corpus statistics, as in the ChatGPT \citep{singh2025openai}, Qwen \citep{bai2023qwen,yang2025qwen3}, DeepSeek \citep{deepseek2024deepseek,liu2024deepseek} families.

Prior work has shown that released BPE vocabularies can reveal corpus information\footnote{Tokenizer-training corpora may differ from pretraining corpora, but still reflect composition choices of developers.}. \citet{hayase2024data} recover corpus mixture from BPE merge rules, but it remains coarse-grained rather than estimating the ratio of each token. \citet{zhang-etal-2025-speculating} use token IDs to speculate about the prevalence of polluted Chinese tokens, but their setting targets a specific token group rather than general tokens. This motivates our central question: 

\textbf{\textit{Can released tokenizer vocabularies support general token-level estimation of hidden corpora?}}

\begin{figure}[t]
    \centering
    \includegraphics[width=0.97\columnwidth]{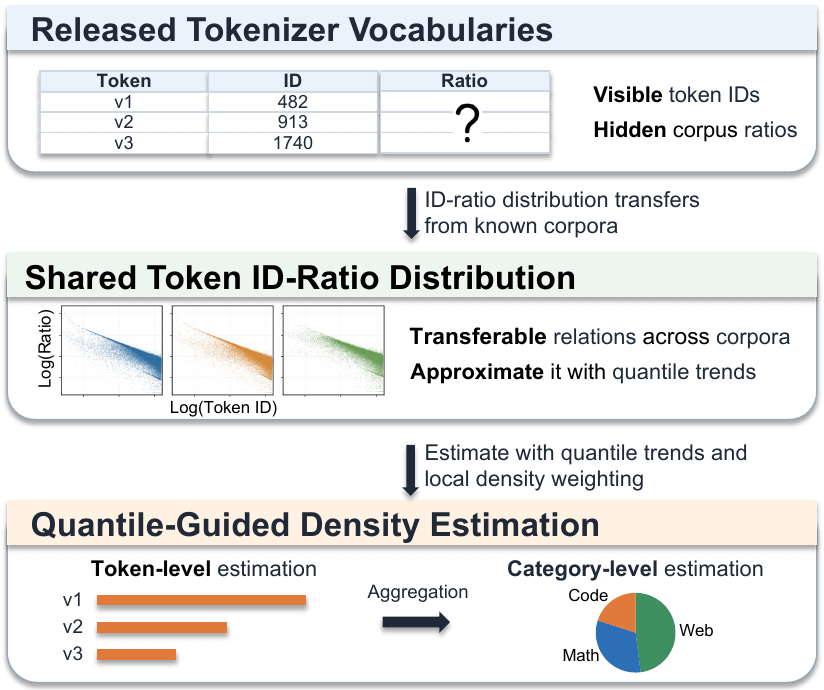}
    \caption{Overview of our work. QGDE transfers ID-ratio structure from known corpora to estimate token and category ratios of hidden corpora.}
    \label{fig:intro_core_idea}
    \vspace{-3ex}
\end{figure}

\autoref{fig:intro_core_idea} illustrates our work. To estimate token ratios, we first compare BPE vocabularies trained on different corpora and find that their token ID--ratio distributions share a stable global shape. This motivates transferring this distribution from known corpora to a target tokenizer trained on hidden corpora. We then propose \textit{Quantile-Guided Density Estimation} (QGDE), which fits multiple quantile trends to approximate the known ID--ratio distribution and enables token-level estimates for the target tokenizer. Our main contributions are:
\begin{packeditemize}
    \item We show that token ID--ratio distributions are transferable across BPE vocabularies trained on different languages and domains, providing a usable signal for estimating hidden corpus ratios from released vocabularies.
    \item We propose QGDE, a general token-level estimator that fits multiple quantile trends to approximate the transferable distribution and uses local density weighting to predict corpus ratios for arbitrary target tokens.
    \item In both controlled settings and a realistic setting, QGDE achieves relative errors as low as 3.00\% for token-level estimation and 3.08\% for category-level estimation after aggregation, surpassing three compared methods.
\end{packeditemize}

\section{Background}


\subsection{Related Work}

\paragraph{Training data transparency and tokenizer signals.}
LLMs are trained on corpora whose exact composition is often undisclosed, making it difficult to interpret the data sources, language coverage, and domain coverage behind released models \citep{bommasani2023foundation,longpre2023data,dodge2021documenting,xu2024benchmark}. Existing transparency methods often rely on model outputs or query access to reveal memorized examples, extract training data, or detect benchmark overlap \citep{carlini2021extracting,carlini2023quantifying,yang2023rethinking}. In contrast, released tokenizers provide a corpus signal because they are learned from corpus statistics: frequent adjacent token pairs are merged earlier, and the resulting vocabulary and token IDs can encode distributional traces of the tokenizer-training corpus \citep{sennrich-etal-2016-neural,hayase2024data,zhang-etal-2025-speculating}.

\paragraph{From coarse mixture to token-level estimation.}
Data Mixture Inference (DMI) \citep{hayase2024data} treats released BPE merge rules as a composition signal: it compares tokenizer merge statistics with candidate corpora and estimates a small vector of category proportions. This output is inherently coarse-grained, over languages, domains, or data sources, rather than a ratio estimate for each token. PoCTrace \citep{zhang-etal-2025-speculating} exploits the relation between token IDs and frequency: it fits a single ID--ratio median trend to estimate the prevalence of polluted Chinese tokens. However, its use case remains tied to a specific token group and a coarse trend estimate. We instead study general token-level corpus ratio estimation for arbitrary tokens.

\subsection{Problem Formulation}

Let \(\mathcal{V}^{*}=\{(v_i,t_i)\}_{i=1}^{N}\) denote the released vocabulary of a target tokenizer, where \(v_i\) is a token string and \(t_i\) is its token ID. The tokenizer-training corpus \(\mathcal{C}^{*}\) is hidden. Let \(r_i^{*}\) denote the corpus ratio of \(v_i\) in \(\mathcal{C}^{*}\). The hidden corpus therefore induces an unknown ID-to-ratio mapping \(f^{*}(t_i)=r_i^{*}\).

We assume access to one or more \textit{known corpora} \(\mathcal{D}\). For each known corpus or mixture of known corpora, token ratios are observable after training and applying a BPE tokenizer \citep{sennrich-etal-2016-neural,hayase2024data}. These known corpora provide empirical ID--ratio pairs that can be used to approximate the unknown mapping \(f^{*}\).

Our goal is to learn an estimator \(\widehat{f}\) from the known ID--ratio pairs, such that for arbitrary tokens in the released target vocabulary,
\begin{equation}
\widehat{f}(t_i)=\widehat{r}_i\approx f^{*}(t_i)=r_i^{*}.
\end{equation}

After estimating token-level ratios, we aggregate them over any token set \(S\subseteq\mathcal{V}^{*}\), e.g., \(\widehat{R}(S)=\sum_{i\in S}\widehat{r}_i\). When token sets are defined by languages or domains, this yields corpus mixture estimates.

\section{Is ID-Ratio Relationship Transferable?}
\label{sec:id_ratio_transferability}

The formulation above relies on a basic assumption: ID--ratio pairs observed in known corpora should provide useful evidence for the hidden target corpus. Since BPE token IDs reflect merge order and thus corpus statistics \citep{sennrich-etal-2016-neural,hayase2024data,zhang-etal-2025-speculating}, the key question is \textbf{\textit{whether the ID--ratio relationship is transferable across tokenizers trained on different corpora}}.

\begin{figure*}[t]
    \centering
    \includegraphics[width=0.95\textwidth]{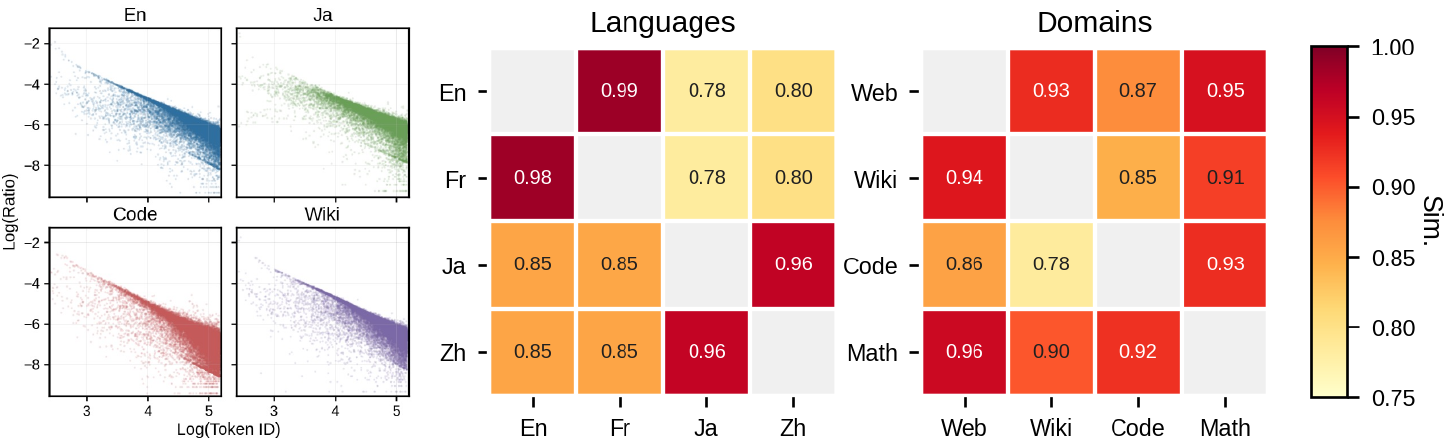}
    \vspace{-0.5em}
    \caption{Token ID--ratio distributions are transferable across different corpora. \textbf{Left}: ID--ratio scatter plots in log--log space. \textbf{Middle}: Similarity across different languages. \textbf{Right}: Similarity across different domain.}
    \label{fig:transfer}
    \vspace{-1em}
\end{figure*}

We compare ID--ratio distributions under two controlled settings: \textbf{\textit{same domain but different languages, and same language but different domains}}. Specifically, we train BPE tokenizers on English, French, Japanese, and Chinese slices from mC4 \citep{xue2021mt5} for the language setting, and on FineWeb (Web) \citep{penedo2024fineweb}, Wikipedia (Wiki) \citep{wikimedia2023wikipedia}, CodeParrot (Code) \citep{codeparrot2021}, and OpenWebMath (Math) \citep{paster2023openwebmath} for the domain setting. For each tokenizer, we remove the initial vocabulary tokens, since special tokens and alphabet symbols do not reflect BPE merge order. We then compute each token's ratio as its count divided by the total token count. 

We plot the token ID--ratio distributions in log--log space in \autoref{fig:transfer}. The scatter plots show that \textit{the distributions are visually similar}, even for English and Japanese, and for Code and Wiki, which differ substantially in language or domain structure (see \autoref{fig:appendix_all_scatter} for all eight corpora). This suggests a transferable ID--ratio distribution.

To quantify this similarity, we convert each scatter plot into a two-dimensional probability density over $\left(\log \mathrm{ID}, \log \mathrm{ratio}\right)$ (see \autoref{app:transfer_similarity}) and define a directional transfer similarity score. Let $P_S$ and $P_T$ denote the ID--ratio densities of a source tokenizer $S$ and a target tokenizer $T$. We compute
\begin{equation}
\mathrm{Sim}(S\to T)=
\exp\left(
-\frac{D_{\mathrm{KL}}(P_T\|P_S)}{H(P_T)}
\right),
\end{equation}
where $H(P_T)$ is the entropy of the target density and \(D_{\mathrm{KL}}\) denotes Kullback--Leibler divergence \citep{murphy2022probabilistic}. This score normalizes the transfer divergence by the target entropy and maps it to $(0,1]$, where larger values indicate that the source distribution better explains the target distribution.

The heatmap in \autoref{fig:transfer} shows that \textbf{\textit{token ID--ratio distributions are broadly transferable}}. In the language setting, transfer is strongest between languages with more similar linguistic structures: English and French are nearly interchangeable, and Japanese and Chinese are also highly similar. Transfer across these groups is weaker, but still substantial, with similarities around 0.8 in the harder English/French-to-Japanese/Chinese directions. In the domain setting, all pairs remain highly similar, with the largest gap appearing between Code and Wiki, mirroring the larger structural gap between programming language and encyclopedic text. Additional single- and mixed-source transfer profiles are reported in \autoref{app:mix_transfer_similarity}, and transfer across tokenizer families is examined in \autoref{app:tokenizer-family}. Overall, the ID--ratio relationship has a stable global shape, motivating an estimator that exploits the shared global ID--ratio trend across corpora.

\section{Quantile-Guided Density Estimation}

Since token ID--ratio distributions share a global shape, we estimate token ratios by transferring this distributional structure from known corpora with observed token ratios. Such transfer is inevitably imperfect: \textbf{\textit{the key is to preserve the shared ID--ratio distribution}}.

\begin{figure*}[t]
    \centering
    \includegraphics[width=\textwidth]{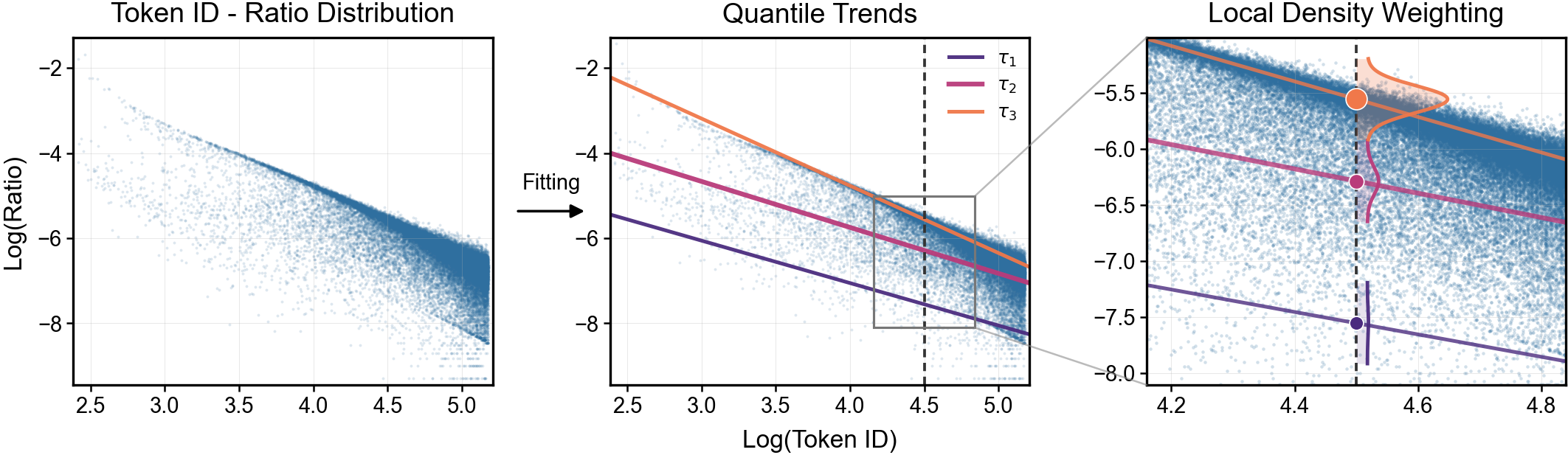}
    \vspace{-1.5em}
    \caption{Overview of quantile-guided density estimation. \textbf{Left}: the ID--ratio distribution in log--log space. \textbf{Middle}: multiple quantile trends approximate the global ID--ratio distribution and produce candidate estimates at a target token ID. \textbf{Right}: Gaussian density kernels around each candidate assign local weights to the estimates.}
    \label{fig:quantile_density_overview}
    \vspace{-0.75em}
\end{figure*}

Inspired by prior work \citep{zhang-etal-2025-speculating} that uses quantile regression to obtain ratio ranges from a single median or boundary curve, we fit multiple quantile trends to better approximate the shared ID--ratio distribution. We then use local density weighting to convert these quantile trends into a token point estimate, as illustrated in \autoref{fig:quantile_density_overview}.

\subsection{Fitting ID--Ratio Trends with Quantiles}
\label{sec:fitting_quantile_trends}

Following Zipf's law \citep{piantadosi2014zipf,saichev2009theory}, which states that word frequency is approximately inversely proportional to frequency rank, a log--log transformation turns this inverse relationship into an approximately linear trend. We therefore model token ID and corpus ratio in log--log space, where quantile regression can fit linear trends at different quantile levels. 

Given a tokenizer trained on known corpora with observed token counts, we represent each token \(j\) as \((x_j,y_j)=(\log t_j,\log r_j)\), where \(t_j\) is the token ID and \(r_j\) is the token's corpus ratio. A single median curve, corresponding to the 0.5 quantile, can capture the central tendency of this relation. However, \textit{it reduces tokens with similar IDs to a single typical ratio, discarding their local ratio variation. This local variation is exactly the distributional signal needed for token-level estimation}.

We instead model the shared ID--ratio relationship \textbf{\textit{with a family of quantile trends}}. Let \(\mathcal{T}\) denote a set of candidate quantile levels. For each \(\tau\in\mathcal{T}\), we fit a log-linear quantile curve over the known ID--ratio points using standard quantile regression \citep{regression2017handbook}, \(q_\tau(x) = a_{\tau}+b_{\tau}x\) (see details in \autoref{app:quantile_fitting}). For a target token with ID \(t_i\), the \(\tau\)-th trend gives a candidate log-ratio estimate:
\begin{equation}
z_{i,\tau}
=
q_\tau(\log t_i)
=
a_\tau+b_\tau\log t_i .
\end{equation}
The quantile family generalizes single-curve token-ID estimators: lower quantiles describe conservative low-ratio hypotheses, upper quantiles describe high-ratio hypotheses, and intermediate quantiles describe the dense central region. Thus, at each token ID, a dense quantile family defines multiple plausible candidate estimates. The remaining question is which quantile levels best approximate the shared ID--ratio distribution.

\subsection{Selecting Quantile Anchors}

To avoid redundant or poorly supported trends, we select a small set of representative quantile anchors
\(\mathcal{T}_K^\star=\{\tau_1,\ldots,\tau_K\}\).
\textit{These anchors should cover the global ID--ratio distribution without collapsing onto redundant regions or passing through consistently sparse regions.} We therefore select anchors by directly maximizing their coverage over the known ID--ratio points \(\mathcal{P}=\{(x_j,y_j)\}_{j=1}^{n}\). 

For a candidate anchor set \(\mathcal{T}_K\), a point is covered if it falls within a vertical band of width \(h_y\) around at least one selected quantile trend. We define Quantile Anchor Coverage \(C\) as
\begin{equation}
C(\mathcal{T}_K)
=
\sum_{(x_j,y_j)\in\mathcal{P}}
\mathbf{1}\!\left[
\min_{\tau\in\mathcal{T}_K}
|y_j-q_\tau(x_j)|<h_y
\right].
\end{equation}
This score counts each known ID--ratio point at most once, even if it is close to multiple selected trends, so redundant anchors receive little extra benefit. We choose the anchor set by maximizing this Quantile Anchor Coverage:
\begin{equation}
\mathcal{T}_K^\star
=
\arg\max_{\mathcal{T}_K\subset\mathcal{T},\,|\mathcal{T}_K|=K}
C(\mathcal{T}_K).
\label{eq:quantile_anchor_selection}
\end{equation}
In implementation, we perform a grid search over quantile levels. Because the maximization depends on both the selected quantile set and the number of anchors \(K\), \autoref{sec:quantile_number_qac} analyzes which anchors are selected and how many anchors are needed.

\subsection{Local Density Weighting}

After selecting quantile anchors, each target token has multiple candidate estimates \(\{z_{i,\tau}\}_{\tau\in\mathcal{T}_K^\star}\). These candidates are plausible under the global ID--ratio trends, but the global trends do not determine how much each candidate should contribute to a particular token. Therefore, \textit{we assign them soft weights by measuring how much local density support each candidate receives from nearby known ID--ratio points}.

For target token \(t_i\), we collect nearby known ID--ratio points \(\mathcal{N}_i=\{(x_j,y_j)\mid |x_j-\log t_i|<h_x\}\). Within this local ID window, we compute the unnormalized local support using Gaussian kernel \citep{silverman1986density}:
\begin{equation}
W_{i,\tau}
=
\sum_{(x_j,y_j)\in \mathcal{N}_i}
\exp\left(
-\frac{(y_j-z_{i,\tau})^2}{2h_y^2}
\right).
\end{equation}
This kernel is a soft version of the vertical neighborhood used in anchor coverage, centered at the candidate estimates \(z_{i,\tau}\).
This weighting matches the intuition in \autoref{fig:quantile_density_overview}: at a fixed token ID, Gaussian density kernels compare how strongly the surrounding points support each quantile candidate.

By normalizing the weights across all quantile trends, the estimated ratio of token \(t_i\) is the density-weighted average of the candidate estimates:
\begin{equation}
\widehat{y}_i
=
\sum_{\tau \in \mathcal{T}_K^\star}
\frac{W_{i,\tau}}
{\sum_{\tau' \in \mathcal{T}_K^\star} W_{i,\tau'}}
z_{i,\tau}.
\end{equation}
Thus, the global quantile trends provide candidate estimates, and local density weighting combines them using token-specific neighborhood evidence. As a result, it turns the range-style signal used in prior token-ID frequency estimation \citep{zhang-etal-2025-speculating} into a fine-grained token-level point estimate.

\section{Quantile Anchor Configuration}
\label{sec:quantile_number_qac}

In \autoref{eq:quantile_anchor_selection}, we select quantile anchors to cover the known ID--ratio distribution while avoiding redundant or poorly supported trends. Before evaluating token-level estimates, we first examine how this Quantile Anchor Coverage (QAC) objective configures the anchors in practice. The following subsections address two practical questions.

\subsection{Which Anchors Cover Better?}
\begin{figure}[t]
    \centering
    \includegraphics[width=0.95\columnwidth]{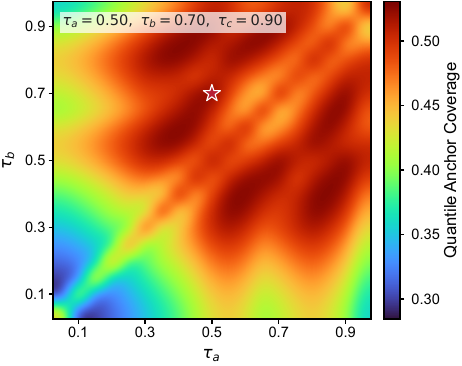}
    \vspace{-1ex}
    \caption{Pairwise projection of Quantile Anchor Coverage for selecting three quantile anchors. The star marks the selected triplet \((0.50,0.70,0.90)\).}
    \label{fig:quantile_anchor_qac}
    \vspace{-0.75em}
\end{figure}

We first inspect the fixed-\(K\) selection problem. Taking \(K=3\) as an example, we grid search over candidate triplets using the QAC objective in \autoref{eq:quantile_anchor_selection}. Since the triplet search space is three-dimensional, \autoref{fig:quantile_anchor_qac} visualizes a pairwise projection: each cell fixes two anchors \((\tau_a,\tau_b)\) and reports the best coverage obtained by any triplet containing that pair.

\autoref{fig:quantile_anchor_qac} shows that coverage is substantially lower when anchors concentrate in sparse or redundant regions. By contrast, the high-coverage area lies around middle-to-high quantile levels. In this \(K=3\) configuration, the triplet \((0.50,0.70,0.90)\) achieves the highest coverage, covering \(53.1\%\) of the sampled known ID--ratio points under the chosen vertical bandwidth. This indicates that QAC does not simply spread anchors uniformly across quantile levels; instead, it favors anchors whose trends jointly pass through well-supported regions of the known ID--ratio distribution.

\subsection{How Many Anchors Are Enough?}
\begin{figure}[t]
    \centering
    \includegraphics[width=0.99\columnwidth]{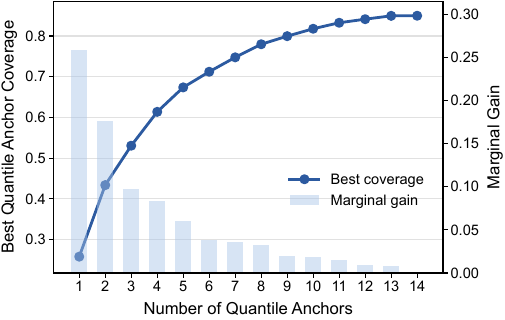}
    \caption{Effect of the number of quantile anchors on Quantile Anchor Coverage. Coverage increases with \(K\), but marginal gains quickly saturate.}
    \label{fig:quantile_number_qac}
    \vspace{-0.75em}
\end{figure}

We then study how the number of quantile anchors affects coverage. For each anchor number \(K\), we maximize \(C(\mathcal{T}_K)\) and report both the best coverage and the marginal gain over \(K-1\) anchors in \autoref{fig:quantile_number_qac}. Coverage increases as more anchors are added, but the marginal gain decreases rapidly and becomes nearly zero at \(K=14\). This saturation suggests that additional anchors eventually pass through regions that are already covered by the selected trends, rather than adding substantial new support from the known ID--ratio distribution. 

Overall, QAC favors complementary anchors rather than sparse or redundant ones. Its coverage then saturates as \(K\) grows, indicating diminishing returns from adding more anchors. A target-independent held-out procedure for selecting \(K\) is reported in \autoref{app:heldout-k}.

\begin{table*}[t]
\centering
\scriptsize
\setlength{\tabcolsep}{2pt}
\renewcommand{\arraystretch}{0.88}
\resizebox{\textwidth}{!}{%
\begin{tabular}{@{}l@{\hspace{5pt}}cccc@{\hspace{3pt}}c@{\hspace{7pt}}cccc@{\hspace{3pt}}c@{\hspace{7pt}}c@{}}
\toprule
\multirow{2}{*}{Source} & \multicolumn{5}{c}{Single-source} & \multicolumn{5}{c}{70\%-mixed source} & \multicolumn{1}{c}{Target-like} \\
\cmidrule(lr){2-6}\cmidrule(lr){7-11}\cmidrule(l){12-12}
& \(S_1\)-only & \(S_2\)-only & \(S_3\)-only & \(S_4\)-only & \makecell{Avg.\(\pm\)Std.} & \(S_1\)-major & \(S_2\)-major & \(S_3\)-major & \(S_4\)-major & \makecell{Avg.\(\pm\)Std.} & Uniform \\
\midrule
\rowcolor{gray!12}
\multicolumn{12}{c}{\textbf{Language Sources} (\(S_1=\)En, \(S_2=\)Fr, \(S_3=\)Zh, \(S_4=\)Ja)} \\
\midrule
Transfer & 21.43 & 21.40 & 31.25 & 37.71 & \avgstd{27.95}{6.92} & 22.19 & 19.58 & 33.53 & 22.33 & \avgstd{24.41}{5.38} & 21.72 \\
PoCTrace & 24.94 & 25.01 & 10.33 & 9.40 & \avgstd{17.42}{7.56} & 18.82 & 14.21 & 8.27 & 9.75 & \avgstd{12.76}{4.12} & 10.45 \\
\midrule
\textbf{QGDE Avg.} & \textbf{5.80} & \textbf{5.84} & \textbf{6.33} & \textbf{5.21} & \textbf{\avgstd{5.80}{0.40}} & \textbf{5.93} & \textbf{3.91} & \textbf{4.65} & \textbf{4.64} & \textbf{\avgstd{4.78}{0.73}} & \textbf{4.68} \\
\quad {\scriptsize \(K=3\)} & 14.11 & 18.05 & 9.17 & 10.95 & \errhigh{\avgstd{13.07}{3.38}} & 17.54 & 10.02 & 7.29 & 10.78 & \errhigh{\avgstd{11.41}{3.77}} & \errhigh{11.42} \\
\quad {\scriptsize \(K=4\)} & 14.19 & 11.95 & 7.98 & 7.24 & \errhigh{\avgstd{10.34}{2.85}} & 9.65 & 5.79 & 6.57 & 6.76 & \errmid{\avgstd{7.19}{1.46}} & \errmid{7.29} \\
\quad {\scriptsize \(K=5\)} & 7.46 & 7.91 & 7.99 & 6.08 & \errmid{\avgstd{7.36}{0.76}} & 7.55 & 3.47 & 5.90 & 5.68 & \errmid{\avgstd{5.65}{1.45}} & \errmid{5.75} \\
\quad {\scriptsize \(K=6\)} & 4.48 & 4.44 & 7.08 & 4.65 & \errlow{\avgstd{5.16}{1.11}} & 8.36 & 4.61 & 4.55 & 4.18 & \errmid{\avgstd{5.43}{1.70}} & \errlow{3.94} \\
\quad {\scriptsize \(K=7\)} & 3.46 & 3.44 & 6.27 & 5.03 & \errlow{\avgstd{4.55}{1.19}} & 4.24 & 3.26 & 4.90 & 4.47 & \errlow{\avgstd{4.22}{0.60}} & \errlow{4.44} \\
\quad {\scriptsize \(K=8\)} & 4.69 & 4.20 & 5.49 & 4.95 & \errlow{\avgstd{4.83}{0.46}} & 3.83 & 2.92 & 4.03 & 3.40 & \errbest{\avgstd{3.54}{0.43}} & \errbest{3.45} \\
\quad {\scriptsize \(K=9\)} & 3.95 & 3.76 & 5.63 & 4.03 & \errlow{\avgstd{4.34}{0.75}} & 3.03 & 2.79 & 3.89 & 3.50 & \errbest{\avgstd{3.31}{0.42}} & \errbest{3.39} \\
\quad {\scriptsize \(K=10\)} & 3.85 & 3.14 & 4.84 & 4.07 & \errbest{\avgstd{3.98}{0.61}} & 3.53 & 2.79 & 3.39 & 3.52 & \errbest{\avgstd{3.31}{0.31}} & \errbest{3.51} \\
\quad {\scriptsize \(K=11\)} & 3.20 & 3.17 & 5.10 & 3.54 & \errbest{\avgstd{3.75}{0.79}} & 3.57 & 2.82 & 3.51 & 3.09 & \errbest{\avgstd{3.25}{0.31}} & \errbest{3.00} \\
\quad {\scriptsize \(K=12\)} & 2.91 & 2.94 & 5.29 & 3.76 & \errbest{\avgstd{3.72}{0.97}} & 3.09 & 2.77 & 3.71 & 3.22 & \errbest{\avgstd{3.20}{0.34}} & \errbest{3.09} \\
\quad {\scriptsize \(K=13\)} & 3.54 & 3.43 & 5.49 & 3.98 & \errbest{\avgstd{4.11}{0.82}} & 3.26 & 2.76 & 3.90 & 3.44 & \errbest{\avgstd{3.34}{0.41}} & \errbest{3.28} \\
\quad {\scriptsize \(K=14\)} & 3.78 & 3.66 & 5.64 & 4.23 & \errlow{\avgstd{4.33}{0.79}} & 3.50 & 2.93 & 4.09 & 3.67 & \errbest{\avgstd{3.55}{0.42}} & \errbest{3.55} \\
\midrule
\rowcolor{gray!12}
\multicolumn{12}{c}{\textbf{Domain Sources} (\(S_1=\)Web, \(S_2=\)Wiki, \(S_3=\)Math, \(S_4=\)Code)} \\
\midrule
Transfer & 13.76 & 13.75 & 53.02 & 12.78 & \avgstd{23.33}{17.15} & 10.42 & 17.59 & 20.96 & 9.97 & \avgstd{14.73}{4.70} & 14.19 \\
PoCTrace & 30.41 & 27.50 & 37.72 & 21.13 & \avgstd{29.19}{5.96} & 27.51 & 26.34 & 25.22 & 22.20 & \avgstd{25.32}{1.97} & 24.32 \\
\midrule
\textbf{QGDE Avg.} & \textbf{11.45} & \textbf{8.06} & \textbf{16.77} & \textbf{24.28} & \textbf{\avgstd{15.14}{6.12}} & \textbf{7.34} & \textbf{6.31} & \textbf{8.71} & \textbf{11.28} & \textbf{\avgstd{8.41}{1.86}} & \textbf{7.44} \\
\quad {\scriptsize \(K=3\)} & 29.08 & 18.66 & 28.44 & 31.44 & \errhigh{\avgstd{26.91}{4.89}} & 22.21 & 17.38 & 24.22 & 23.79 & \errhigh{\avgstd{21.90}{2.71}} & \errhigh{22.24} \\
\quad {\scriptsize \(K=4\)} & 25.12 & 17.29 & 27.43 & 32.78 & \errhigh{\avgstd{25.66}{5.57}} & 19.59 & 10.85 & 21.70 & 22.15 & \errhigh{\avgstd{18.57}{4.56}} & \errmid{11.89} \\
\quad {\scriptsize \(K=5\)} & 17.87 & 10.45 & 26.06 & 31.42 & \errhigh{\avgstd{21.45}{7.97}} & 4.66 & 4.89 & 14.52 & 17.99 & \errmid{\avgstd{10.52}{5.87}} & \errmid{12.76} \\
\quad {\scriptsize \(K=6\)} & 18.60 & 11.22 & 23.76 & 30.94 & \errhigh{\avgstd{21.13}{7.21}} & 5.91 & 4.85 & 7.64 & 15.04 & \errmid{\avgstd{8.36}{3.98}} & \errlow{5.42} \\
\quad {\scriptsize \(K=7\)} & 13.19 & 7.03 & 20.64 & 31.67 & \errmid{\avgstd{18.13}{9.18}} & 4.47 & 4.48 & 4.73 & 10.96 & \errlow{\avgstd{6.16}{2.78}} & \errbest{4.44} \\
\quad {\scriptsize \(K=8\)} & 5.33 & 4.52 & 20.43 & 31.81 & \errmid{\avgstd{15.52}{11.34}} & 4.52 & 4.69 & 4.67 & 11.45 & \errlow{\avgstd{6.33}{2.96}} & \errbest{4.44} \\
\quad {\scriptsize \(K=9\)} & 5.15 & 4.50 & 16.77 & 31.28 & \errmid{\avgstd{14.43}{10.89}} & 4.48 & 4.89 & 4.47 & 11.02 & \errlow{\avgstd{6.21}{2.78}} & \errbest{4.52} \\
\quad {\scriptsize \(K=10\)} & 4.49 & 4.66 & 8.63 & 30.05 & \errmid{\avgstd{11.96}{10.58}} & 4.45 & 4.84 & 4.45 & 5.00 & \errbest{\avgstd{4.69}{0.24}} & \errbest{4.76} \\
\quad {\scriptsize \(K=11\)} & 4.70 & 4.59 & 8.81 & 10.63 & \errlow{\avgstd{7.18}{2.62}} & 4.49 & 5.01 & 4.58 & 4.49 & \errbest{\avgstd{4.64}{0.22}} & \errbest{4.99} \\
\quad {\scriptsize \(K=12\)} & 4.54 & 4.67 & 7.32 & 10.53 & \errlow{\avgstd{6.76}{2.44}} & 4.45 & 4.71 & 4.46 & 4.45 & \errbest{\avgstd{4.52}{0.11}} & \errbest{4.67} \\
\quad {\scriptsize \(K=13\)} & 4.63 & 4.57 & 7.61 & 9.54 & \errlow{\avgstd{6.59}{2.10}} & 4.45 & 4.60 & 4.51 & 4.51 & \errbest{\avgstd{4.52}{0.05}} & \errbest{4.60} \\
\quad {\scriptsize \(K=14\)} & 4.65 & 4.52 & 5.30 & 9.27 & \errlow{\avgstd{5.94}{1.95}} & 4.46 & 4.56 & 4.53 & 4.51 & \errbest{\avgstd{4.52}{0.04}} & \errbest{4.54} \\
\bottomrule
\end{tabular}}
\caption{Mean relative error (MRE) (\%) of token-level ratio estimation across language and domain source mixtures. Single-source uses one known corpus; 70\%-mixed uses a 70\%-dominant known-corpus mixture, with the remaining three sources mixed equally; Target-like uses the same uniform mixture as the target corpora. Darker green indicates higher error in the QGDE \(K\)-ablation rows.}
\label{tab:token_level_estimation}
\vspace{-1em}
\end{table*}

\section{Token-Level Ratio Estimation}

We now test whether QGDE turns the quantile trends into accurate token-level ratio estimates.

\subsection{Evaluation Setting}

The evaluation is under different source compositions. Here, source refers to the known corpus mixture used to fit the estimator, while the target corpora are held fixed as uniform mixtures and used only for evaluation.

We use the same language and domain categories as in \autoref{sec:id_ratio_transferability}, but evaluate on target corpora drawn from different datasets. For languages, mC4 serves as the source side and OSCAR \citep{ortizsuarez2020monolingual} as the target side over English, French, Japanese, and Chinese. For domains, Web, Wiki, Code, and Math are paired with target corpora from the same broad domains: RedPajama-C4 \citep{weber2024redpajama}, BookCorpus \citep{zhu2015aligning}, RedPajama-GitHub \citep{weber2024redpajama}, and FineWebMath \citep{allal2025smollm2}.

We report mean relative error in \autoref{tab:token_level_estimation}, where lower values indicate better estimates. We compare QGDE with two baselines: direct ID-ratio transfer, which copies the source ratio profile by token position, and PoCTrace \citep{zhang-etal-2025-speculating}, which estimates ratios from a single median ID--ratio trend.

\subsection{Token-Level Results}

\autoref{tab:token_level_estimation} yields four main observations.

\paragraph{QGDE outperforms baselines.}
QGDE consistently improves token-level estimation over both baselines. Direct ID-ratio transfer copies the source ratio profile and therefore remains high-error, showing that distributional transferability does not justify token-wise ratio copying. PoCTrace avoids direct copying by fitting a median ID--ratio trend, but a single trend cannot represent the vertical spread of plausible ratios at each token ID. In contrast, QGDE combines multiple quantile trends with local density weighting. The later \(K\)-rows are substantially lower than both baselines in most source-mixture settings, especially for mixed known-corpus settings.


\paragraph{More quantile anchors help, then saturate.}
The ablation on \(K\) confirms that multiple quantile anchors are necessary, but that their benefit saturates. Moving from \(K=3\) to larger \(K\) sharply reduces error, especially in the domain block: the single-source average decreases from 26.91 to 5.94, and the 70\%-mixed average decreases from 21.90 to 4.52. Later rows fluctuate within a narrower band, so the best \(K\) is not universal across source compositions; nevertheless, \(K=14\) provides a reasonable high-coverage default once the QAC gain has saturated. This echoes the QAC analysis in \autoref{sec:quantile_number_qac}: once additional anchors provide little new coverage of the known ID--ratio distribution, they also yield diminishing improvements in token-level estimation error.

\paragraph{Mixed sources help, but exact ratios matter less.}
Mixed sources are generally preferable to single-source settings because they expose the estimator to a broader ID--ratio distribution. This is most visible in the domain block, where the QGDE average drops from 15.14 for single-source settings to 8.41 for 70\%-mixed settings. Within the mixed-source group, however, the exact dominant component matters much less: at \(K=14\), the 70\%-mixed columns have low standard deviation, with Std. of 0.42 for languages and 0.04 for domains. Even matching the target mixture exactly (Target-like at the last column) is not always optimal; what matters more is using a mixed source that covers multiple source components.

\paragraph{Language ratios are easier to predict than domain ratios.}
The language block reaches low error with fewer anchors, whereas the domain block has much larger errors at small \(K\) and only approaches a similar range after more anchors are used. This difference reflects how closely the source and target ID--ratio relationships match. In the language setting, mC4 and OSCAR differ as corpora, but the language-specific signals that shape token IDs are largely stable across them. In the domain setting, the matched source and target corpora are less aligned. Source and target corpora may differ in collection pipelines, so the same broad domain label does not guarantee a similar ID--ratio relationship.

\section{Aggregating Token Ratios into Mixtures}
\label{sec:aggregating_token_ratios}

A useful token-level estimator should also support corpus mixture estimation. We test this by aggregating QGDE's token-level estimates into language or domain proportions.

\subsection{Aggregation Procedure}
We convert target token ratios into category proportions by distributing each estimated token ratio \(\widehat r_i\) \textit{according to how the token appears across known source categories}. For example, a token that appears mostly in one category contributes most of its estimated ratio to that category, while a token that appears across all categories is split according to its relative counts in the known corpora. Formally, we define the category assignment weight as
\begin{equation}
\pi_{c,i}
=
\frac{n_{c,i}}{\sum_{c'\in\mathcal{C}}n_{c',i}},
\qquad c\in\mathcal{C}.
\end{equation}
Here, \(\mathcal{C}\) is the set of source categories, and \(n_{c,i}\) is the count of target token \(v_i\) in the known corpus for category \(c\).

We then normalize the estimated token ratios over token set \(\mathcal{I}\) and distribute each token ratio to categories using \(\pi_{c,i}\). The estimated mixture proportion of category \(c\) is therefore
\begin{equation}
\widehat{\alpha}_c
=
\sum_{i\in \mathcal{I}}
\frac{\widehat r_i}{\sum_{j\in \mathcal{I}}\widehat r_j}
\pi_{c,i},
\qquad c\in\mathcal{C}.
\end{equation}

\begin{figure*}[t]
    \centering
    \includegraphics[width=\textwidth]{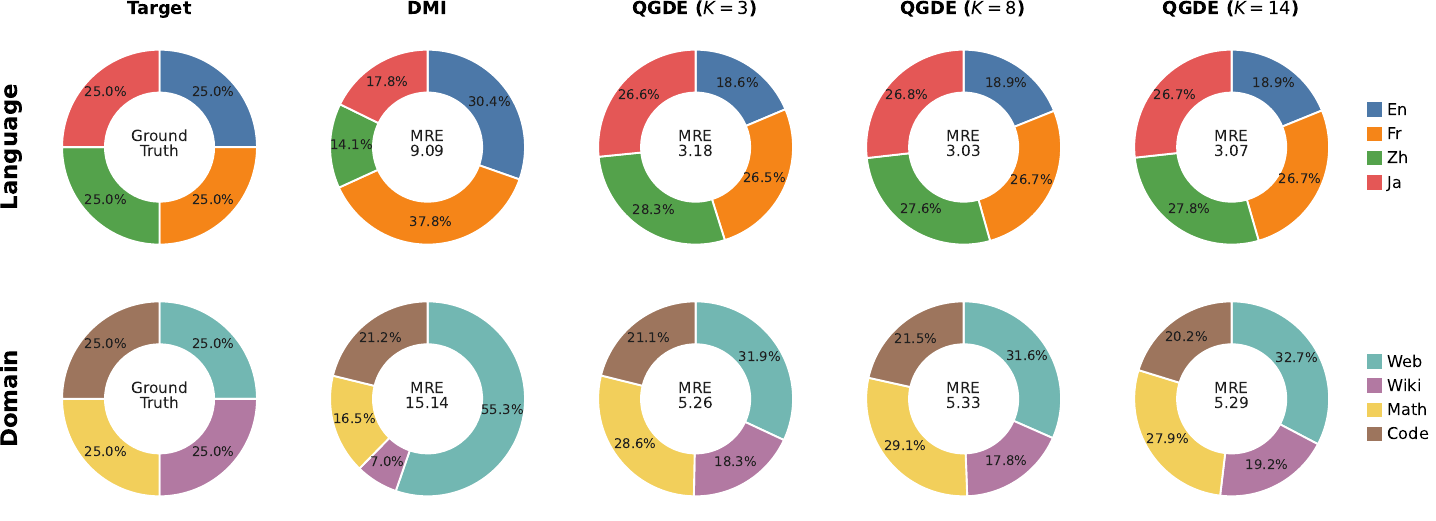}
    \caption{Category-level mixture estimation, where QGDE matches the ground truth more closely in both language and domain settings. Each setting compares the ground truth, DMI, and QGDE at three \(K\) values; center labels report mean relative error (MRE).}
    \label{fig:category_level_recovery}
\end{figure*}

\subsection{Category-Level Results}
\autoref{fig:category_level_recovery} evaluates category-level mixture estimation under uniform language and domain targets. We compare QGDE aggregation with DMI \citep{hayase2024data}, a source-independent baseline that specifically designed for estimating macro mixture proportions from tokenizer statistics.

\paragraph{Token-level estimates support mixture estimation.}
The figure shows that QGDE token-level ratios can be aggregated into meaningful category-level proportions. Across both language and domain targets, the QGDE pies track the ground-truth mixtures closely than the baseline estimates. This indicates that the token-level signal recovered by QGDE is not only useful for individual token prediction, but also remains informative after aggregation to categories.

\paragraph{QGDE outperforms baseline DMI.}
DMI gives visibly skewed estimates: in the language setting, it overestimates English and French while underestimating Chinese and Japanese; in the domain setting, it strongly overestimates Web and underestimates the remaining categories. QGDE substantially reduces these errors, lowering language error from 9.09 to about 3.0 and domain error from 15.14 to a much smaller range.

\paragraph{Anchor gains are weaker after aggregation.}
The effect of increasing \(K\) is less pronounced at the category level than at the token level. In the language setting, error decreases from \(K=3\) to \(K=8\), but changes only slightly at \(K=14\). In the domain setting, the trend is even less monotonic, because aggregation can shift estimated ratios among overlapping domain vocabularies even when token-level predictions improve. Thus, additional anchors still help by improving the underlying token estimates, but their gains are partially smoothed or redistributed by category-level aggregation. The full source-composition and \(K\)-sweep results are reported in \autoref{tab:appendix_category_level_uniform}.

\section{Validation on the SmolLM Tokenizer}

Beyond the controlled language and domain settings, we also test QGDE in a realistic setting: \textit{a released tokenizer along with training corpora}. Such validation is uncommon because LLM releases rarely include training corpora. SmolLM \citep{allal2025smollm2} is a useful exception.

We use the SmolLM tokenizer as the target tokenizer. Token-level ground truth is computed from the training corpus, and category-level ground truth uses the released component proportions: FineWeb-edu 87.30\%, Cosmopedia-v2 11.11\%, and Python-edu 1.59\%. Since the exact component corpora cannot be used as known corpora, we fit the estimators on corpora matched to the three components: RedPajama-C4 \citep{weber2024redpajama}, a Wikipedia--arXiv mixture \citep{wikimedia2023wikipedia,arxiv2020dataset}, and CodeParrot \citep{codeparrot2021}, respectively. We evaluate three known-corpus mixture ratios over these proxies: a uniform mixture, a pretrain-like mixture based on \citep{dolma3}, and a target-like mixture that matches the released SmolLM component proportions as a diagnostic setting. \autoref{tab:smollm_real_world} shows the results.

\begin{table}[t]
\centering
\scriptsize
\renewcommand{\arraystretch}{0.95}
\resizebox{0.9\linewidth}{!}{%
\begin{tabular}{@{}lccc@{}}
\toprule
Source & Uniform & Pretrain-like & Target-like \\
\midrule
\rowcolor{gray!12}
\multicolumn{4}{c}{Token-level} \\
\addlinespace[0.15em]
Transfer & 9.06 & 10.53 & 7.71 \\
PoCTrace & 13.84 & 18.68 & 20.33 \\
\textbf{QGDE} & \textbf{5.78} & \textbf{5.73} & \textbf{5.72} \\
\midrule
\rowcolor{gray!12}
\multicolumn{4}{c}{Category-level} \\
\addlinespace[0.15em]
DMI & 9.11 & 9.11 & 9.11 \\
\textbf{QGDE} & \textbf{6.08} & \textbf{5.90} & \textbf{5.93} \\
\bottomrule
\end{tabular}}
\caption{Validation on the released SmolLM tokenizer, where QGDE achieves the lowest error for both token-level and category-level estimation.}
\label{tab:smollm_real_world}
\vspace{-1ex}
\end{table}

At the token-level, QGDE outperforms both direct ID-ratio transfer and PoCTrace, reaching 5.72--5.78 MRE. The small spread across Uniform, Pretrain-like, and Target-like source mixture also echoes the controlled experiments: once the known-corpus mixture covers the relevant components, the exact mixture ratio is less important. An additional validation of Pythia \citep{biderman2023pythia} is reported in \autoref{app:pythia-validation}.

At the category level, QGDE also improves over the source-independent DMI baseline, reducing error from 9.11 to about 5.9--6.1. These estimates are less exact than in the controlled category-level experiments because the component proportions are highly imbalanced; in particular, Python-edu accounts for only 1.59\%. Nevertheless, QGDE still recovers a better mixture estimate. A direct proxy-corpus replacement experiment is reported in \autoref{app:proxy-replacement}.

\section{Conclusion}

Released LLM vocabularies provide a useful signal for estimating hidden corpus composition beyond coarse category proportions. We show that BPE tokenizers share stable token ID--ratio distributions across corpora, and introduce QGDE to transfer this structure through quantile trends and local density weighting. Across controlled settings and the released SmolLM tokenizer, QGDE achieves relative errors as low as 3.00\% for token-level estimation and 3.08\% after aggregation into category-level mixtures.

\clearpage
\section*{Limitations}
\paragraph{Scarcity of ground truth for released LLM tokenizers.}
ChatGPT, Qwen, and DeepSeek release tokenizer vocabularies, but not their training corpora. This prevents direct evaluation of token-level ratio estimates on these models. We therefore validate QGDE in controlled settings and on SmolLM (also Pythia in \autoref{app:pythia-validation}), rare released tokenizers with available training data.

\paragraph{QGDE is designed for BPE tokenizers.}
As stated in \autoref{sec:intro}, QGDE is designed for BPE tokenizers. The diagnostic in \autoref{app:tokenizer-family} shows weaker transfer to Unigram and WordPiece tokenizers, so applying QGDE unchanged to those constructions is outside the scope. Also, manually concatenated BPE tokenizers is out of scope. Nevertheless, this BPE scope still covers many mainstream released decoder-only LLM tokenizers.

\section*{Ethics Statement}
ACL Ethics Policy is respected in this work. This work studies corpus ratio estimation from released tokenizer vocabularies and known corpora. We use publicly available or controlled corpora for research purposes, and we respect the terms, conditions, and copyright requirements of the corresponding data sources. No human subjects or private personal data are involved. The proposed methods are intended for research use, transparency analysis, and auditing of corpus composition signals from released tokenizers.

We adhere to the Association for Computational Linguistics (ACL) guidelines on responsible NLP research\footnote{\url{https://aclrollingreview.org/responsibleNLPresearch/}}, with particular attention to transparency, research-use framing, and responsible handling of corpus-derived evidence.

\section*{Use of AI Assistants}
The authors used AI assistants for language polishing, LaTeX editing, and phrasing suggestions during paper preparation. All substantive claims, experimental results, analyses, citations, and final text were reviewed and verified by the authors.

\section*{Acknowledgements}
This work was supported by Alibaba Group through Alibaba Innovative Research Program.

\bibliography{custom}

\clearpage
\appendix

\section{Quantile Trend Fitting Details}
\label{app:quantile_fitting}

In \autoref{sec:fitting_quantile_trends}, QGDE fits one log-linear trend for each quantile level over the known ID--ratio points using quantile regression \citep{regression2017handbook}. This section gives the optimization objective used to fit those trends.

Let \(\mathcal{P}=\{(x_j,y_j)\}_{j=1}^{n}\) denote the known ID--ratio points, where \(x_j=\log t_j\), \(y_j=\log r_j\), \(t_j\) is the token ID, and \(r_j\) is the observed ratio in the known corpus. For each quantile level \(\tau\in\mathcal{T}\), we estimate the trend coefficients by
\begin{equation}
(a_\tau,b_\tau)
=
\arg\min_{a,b}
\sum_{j=1}^{n}
\rho_{\tau}\left(y_j-a-bx_j\right),
\end{equation}
where \(\rho_\tau(\cdot)\) is the asymmetric loss:
\begin{equation}
\rho_{\tau}(u)
=
\begin{cases}
\tau u, & u \geq 0, \\
(\tau - 1)u, & u < 0.
\end{cases}
\end{equation}
The fitted trend is \(q_\tau(x)=a_\tau+b_\tau x\). The asymmetric loss encourages approximately a \(\tau\) fraction of known points to lie below the trend, so different \(\tau\) values trace different vertical levels of the ID--ratio distribution. A single median trend captures only the central tendency; QGDE fits a family of trends so that the vertical spread of plausible ratios is preserved before local density weighting.

\section{Transfer Similarity Computation}
\label{app:transfer_similarity}

In \autoref{sec:id_ratio_transferability}, \autoref{fig:transfer} quantifies whether ID--ratio distributions transfer across tokenizers. We compute this score by converting each scatter plot into a probability density \(P_D\) over ID--ratio space.

For each tokenizer \(D\), each token \(v_i\) is represented as \((x_i,y_i)=(\log t_i,\log r_i)\), where \(t_i\) is its token ID and \(r_i\) is its corpus ratio. We place all tokenizers on a shared grid \(\mathcal{B}\) and convert each tokenizer into a smoothed two-dimensional histogram:
\begin{equation}
P_D(b)=
\frac{\sum_{v_i\in V_D}\mathbf{1}\!\left[(x_i,y_i)\in b\right]+\epsilon}
{|V_D|+\epsilon|\mathcal{B}|},
\qquad b\in\mathcal{B}.
\end{equation}
Here, \(\epsilon\) is a small smoothing constant. Setting \(D=S\) or \(D=T\) gives the source and target densities \(P_S\) and \(P_T\) used in the main text.

The score is directional: \(\mathrm{Sim}(S\to T)\) asks how well the source density \(P_S\) explains the target density \(P_T\). We therefore compute \(D_{\mathrm{KL}}(P_T\|P_S)\), normalize it by the target entropy \(H(P_T)\), and map the result to \((0,1]\), where higher values indicate stronger transfer similarity \citep{murphy2022probabilistic}.

\section{Single-source and Mixed-source Transfer Similarity}
\label{app:mix_transfer_similarity}

In \autoref{sec:id_ratio_transferability}, the main transfer analysis compares single-category tokenizers. Because the token-level experiments also use mixed known-corpus sources, \autoref{tab:domain_mix_transfer} and \autoref{tab:language_mix_transfer} report transfer similarity from both single-source and mixed-source tokenizers to single-category targets. Rows specify the source tokenizer's training mixture, and columns specify the target tokenizer whose ID--ratio density is explained.

\begin{table}[t]
\centering
\footnotesize
\begin{tabularx}{\columnwidth}{@{}>{\centering\arraybackslash}p{0.32\columnwidth}*{4}{>{\centering\arraybackslash}X}@{}}
\toprule
\multicolumn{1}{c}{Source Ratio} & \multicolumn{4}{c}{Target} \\
\cmidrule(lr){2-5}
Web:Wiki:Code:Math & Web & Wiki & Code & Math \\
\midrule
100:0:0:0 & 1.00 & 0.93 & 0.87 & 0.95 \\
0:100:0:0 & 0.94 & 1.00 & 0.85 & 0.91 \\
0:0:100:0 & 0.86 & 0.78 & 1.00 & 0.93 \\
0:0:0:100 & 0.96 & 0.90 & 0.92 & 1.00 \\
\midrule
70:10:10:10 & 0.96 & 0.93 & 0.84 & 0.91 \\
10:70:10:10 & 0.93 & 0.97 & 0.82 & 0.89 \\
10:10:70:10 & 0.94 & 0.96 & 0.82 & 0.89 \\
10:10:10:70 & 0.97 & 0.91 & 0.87 & 0.94 \\
25:25:25:25 & 0.95 & 0.95 & 0.83 & 0.91 \\
\bottomrule
\end{tabularx}
\caption{Directional transfer similarity of domains.}
\label{tab:domain_mix_transfer}
\end{table}

\begin{table}[t]
\centering
\footnotesize
\begin{tabularx}{\columnwidth}{@{}>{\centering\arraybackslash}p{0.32\columnwidth}*{4}{>{\centering\arraybackslash}X}@{}}
\toprule
\multicolumn{1}{c}{Source Ratio} & \multicolumn{4}{c}{Target} \\
\cmidrule(lr){2-5}
En:Fr:Ja:Zh & En & Fr & Ja & Zh \\
\midrule
100:0:0:0 & 1.00 & 0.99 & 0.78 & 0.80 \\
0:100:0:0 & 0.99 & 1.00 & 0.78 & 0.80 \\
0:0:100:0 & 0.85 & 0.85 & 1.00 & 0.96 \\
0:0:0:100 & 0.85 & 0.85 & 0.96 & 1.00 \\
\midrule
70:10:10:10 & 0.95 & 0.95 & 0.89 & 0.88 \\
10:70:10:10 & 0.91 & 0.91 & 0.94 & 0.93 \\
10:10:70:10 & 0.84 & 0.84 & 0.97 & 0.96 \\
10:10:10:70 & 0.84 & 0.84 & 0.97 & 0.96 \\
25:25:25:25 & 0.87 & 0.87 & 0.97 & 0.95 \\
\bottomrule
\end{tabularx}
\caption{Directional transfer similarity of languages.}
\label{tab:language_mix_transfer}
\end{table}

The mixed-source profiles support the token-level results in \autoref{tab:token_level_estimation}. First, ID--ratio distributions remain broadly transferable after mixing: most mixed-source similarities are still above 0.8, and many domain similarities are close to or above 0.9. Second, mixed sources retain useful similarity to multiple targets rather than only to a single self-matched target. This helps explain why mixed known-corpus sources are generally more stable for token-level estimation. Third, the same structure as in the main heatmap remains visible: English--French and Japanese--Chinese form stronger language clusters, while Code is the hardest domain target because its ID--ratio distribution differs more from natural language corpora.

\begin{figure*}[t]
    \centering
    \includegraphics[width=0.98\textwidth]{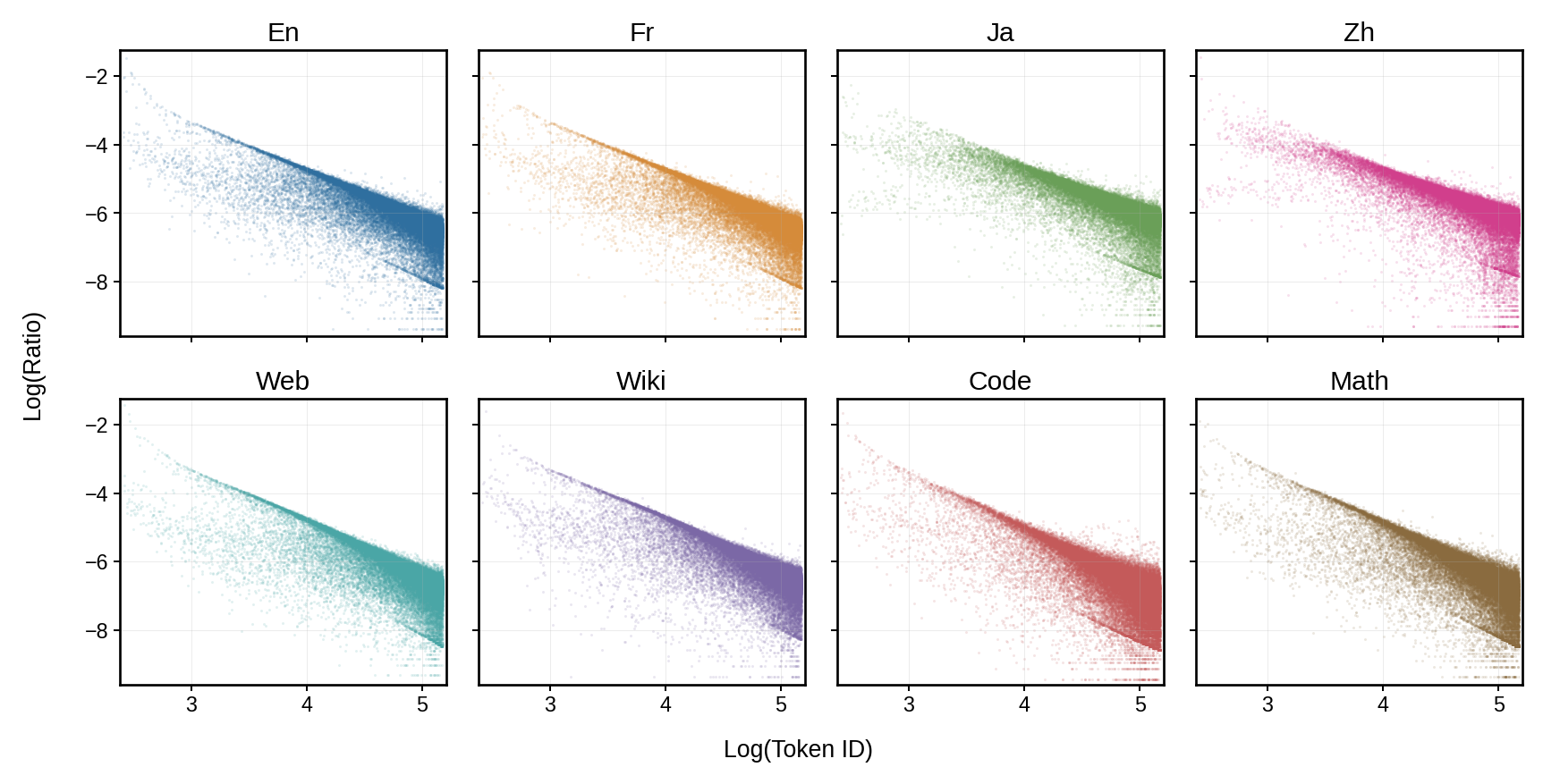}
    \caption{Token ID--ratio scatter plots for all eight controlled tokenizers. The top row shows language-specific mC4 tokenizers, and the bottom row shows domain-specific English tokenizers.}
    \label{fig:appendix_all_scatter}
\end{figure*}

\section{Additional ID--Ratio Scatter Plots}
\label{app:additional_scatter}

In \autoref{sec:id_ratio_transferability}, \autoref{fig:transfer} shows representative ID--ratio scatter plots to motivate transferability. \autoref{fig:appendix_all_scatter} provides the full set of eight controlled tokenizers: the top row shows language-specific mC4 tokenizers, and the bottom row shows domain-specific English tokenizers.

Across both rows, the point clouds follow a similar downward log--log shape, while their thickness, location, and tail behavior vary by language and domain. These patterns support the main-text conclusion that the ID--ratio relationship is broadly shared but not identical across corpora.

\section{Category-Level Estimation Details}
\label{app:category_level_recovery_details}

In \autoref{sec:aggregating_token_ratios}, \autoref{fig:category_level_recovery} shows selected category-level estimates for the uniform language and domain targets. We present the complete source-composition and \(K\)-sweep results in this section, showing that the main conclusions do not depend on a single displayed \(K\) or source mixture.

\begin{table*}[h!]
\centering
\scriptsize
\setlength{\tabcolsep}{2pt}
\renewcommand{\arraystretch}{0.88}
\resizebox{0.98\textwidth}{!}{%
\begin{tabular}{@{}l@{\hspace{5pt}}cccc@{\hspace{3pt}}c@{\hspace{7pt}}cccc@{\hspace{3pt}}c@{\hspace{7pt}}c@{}}
\toprule
\multirow{2}{*}{Source} & \multicolumn{5}{c}{Single-source} & \multicolumn{5}{c}{70\%-mixed source} & \multicolumn{1}{c}{Target-like} \\
\cmidrule(lr){2-6}\cmidrule(lr){7-11}\cmidrule(l){12-12}
& \(S_1\)-only & \(S_2\)-only & \(S_3\)-only & \(S_4\)-only & \makecell{Avg.\(\pm\)Std.} & \(S_1\)-major & \(S_2\)-major & \(S_3\)-major & \(S_4\)-major & \makecell{Avg.\(\pm\)Std.} & Uniform \\
\midrule
\rowcolor{gray!12}
\multicolumn{12}{c}{\textbf{Language Sources} (\(S_1=\)En, \(S_2=\)Fr, \(S_3=\)Zh, \(S_4=\)Ja)} \\
\midrule
DMI & \multicolumn{11}{c}{9.09 \; \textit{(source-independent)}} \\
\midrule
\textbf{QGDE Avg.} & \textbf{3.25} & \textbf{3.21} & \textbf{3.22} & \textbf{3.14} & \textbf{\avgstd{3.21}{0.04}} & \textbf{3.20} & \textbf{3.16} & \textbf{3.14} & \textbf{3.14} & \textbf{\avgstd{3.16}{0.02}} & \textbf{3.12} \\
\quad {\scriptsize \(K=3\)} & 3.70 & 3.58 & 3.33 & 3.43 & \errhigh{\avgstd{3.51}{0.14}} & 3.59 & 3.47 & 3.21 & 3.47 & \errhigh{\avgstd{3.44}{0.14}} & \errhigh{3.42} \\
\quad {\scriptsize \(K=4\)} & 3.62 & 3.53 & 3.26 & 3.23 & \errhigh{\avgstd{3.41}{0.17}} & 3.39 & 3.27 & 3.16 & 3.25 & \errmid{\avgstd{3.27}{0.08}} & \errlow{3.17} \\
\quad {\scriptsize \(K=5\)} & 3.43 & 3.41 & 3.26 & 3.17 & \errmid{\avgstd{3.32}{0.11}} & 3.31 & 3.25 & 3.15 & 3.14 & \errlow{\avgstd{3.22}{0.07}} & \errbest{3.11} \\
\quad {\scriptsize \(K=6\)} & 3.32 & 3.31 & 3.22 & 3.08 & \errmid{\avgstd{3.23}{0.10}} & 3.29 & 3.21 & 3.11 & 3.08 & \errlow{\avgstd{3.17}{0.08}} & \errbest{3.07} \\
\quad {\scriptsize \(K=7\)} & 3.20 & 3.18 & 3.19 & 3.11 & \errlow{\avgstd{3.17}{0.03}} & 3.13 & 3.11 & 3.14 & 3.12 & \errlow{\avgstd{3.12}{0.01}} & \errbest{3.11} \\
\quad {\scriptsize \(K=8\)} & 3.18 & 3.08 & 3.18 & 3.11 & \errlow{\avgstd{3.13}{0.04}} & 3.15 & 3.13 & 3.14 & 3.09 & \errlow{\avgstd{3.13}{0.02}} & \errbest{3.10} \\
\quad {\scriptsize \(K=9\)} & 3.09 & 3.01 & 3.19 & 3.07 & \errbest{\avgstd{3.09}{0.06}} & 3.07 & 3.08 & 3.12 & 3.09 & \errbest{\avgstd{3.09}{0.02}} & \errbest{3.09} \\
\quad {\scriptsize \(K=10\)} & 3.10 & 3.10 & 3.18 & 3.08 & \errbest{\avgstd{3.11}{0.04}} & 3.10 & 3.09 & 3.09 & 3.10 & \errbest{\avgstd{3.09}{0.00}} & \errbest{3.09} \\
\quad {\scriptsize \(K=11\)} & 3.11 & 3.11 & 3.20 & 3.07 & \errlow{\avgstd{3.12}{0.05}} & 3.12 & 3.10 & 3.11 & 3.08 & \errbest{\avgstd{3.10}{0.02}} & \errbest{3.04} \\
\quad {\scriptsize \(K=12\)} & 3.05 & 3.04 & 3.22 & 3.10 & \errbest{\avgstd{3.10}{0.07}} & 3.06 & 3.05 & 3.13 & 3.09 & \errbest{\avgstd{3.08}{0.03}} & \errbest{3.06} \\
\quad {\scriptsize \(K=13\)} & 3.08 & 3.06 & 3.23 & 3.12 & \errlow{\avgstd{3.12}{0.07}} & 3.08 & 3.06 & 3.15 & 3.10 & \errbest{\avgstd{3.10}{0.03}} & \errbest{3.08} \\
\quad {\scriptsize \(K=14\)} & 3.17 & 3.14 & 3.24 & 3.13 & \errlow{\avgstd{3.17}{0.04}} & 3.12 & 3.08 & 3.15 & 3.11 & \errbest{\avgstd{3.12}{0.03}} & \errbest{3.09} \\
\midrule
\rowcolor{gray!12}
\multicolumn{12}{c}{\textbf{Domain Sources} (\(S_1=\)Web, \(S_2=\)Wiki, \(S_3=\)Math, \(S_4=\)Code)} \\
\midrule
DMI & \multicolumn{11}{c}{15.14 \; \textit{(source-independent)}} \\
\midrule
\textbf{QGDE Avg.} & \textbf{5.44} & \textbf{5.37} & \textbf{5.40} & \textbf{5.34} & \textbf{\avgstd{5.39}{0.04}} & \textbf{5.44} & \textbf{5.37} & \textbf{5.48} & \textbf{5.39} & \textbf{\avgstd{5.42}{0.05}} & \textbf{5.37} \\
\quad {\scriptsize \(K=3\)} & 5.25 & 5.44 & 5.27 & 5.24 & \errbest{\avgstd{5.30}{0.08}} & 5.33 & 5.40 & 5.40 & 5.26 & \errlow{\avgstd{5.35}{0.06}} & \errbest{5.32} \\
\quad {\scriptsize \(K=4\)} & 5.54 & 5.26 & 5.39 & 5.27 & \errlow{\avgstd{5.37}{0.11}} & 5.23 & 5.40 & 5.42 & 5.33 & \errlow{\avgstd{5.34}{0.08}} & \errbest{5.28} \\
\quad {\scriptsize \(K=5\)} & 5.50 & 5.35 & 5.29 & 5.27 & \errlow{\avgstd{5.35}{0.09}} & 5.60 & 5.32 & 5.51 & 5.32 & \errhigh{\avgstd{5.44}{0.12}} & \errbest{5.29} \\
\quad {\scriptsize \(K=6\)} & 5.47 & 5.47 & 5.41 & 5.30 & \errhigh{\avgstd{5.42}{0.07}} & 5.50 & 5.36 & 5.46 & 5.27 & \errmid{\avgstd{5.40}{0.09}} & \errhigh{5.43} \\
\quad {\scriptsize \(K=7\)} & 5.51 & 5.39 & 5.35 & 5.34 & \errmid{\avgstd{5.40}{0.07}} & 5.44 & 5.39 & 5.51 & 5.42 & \errhigh{\avgstd{5.44}{0.05}} & \errhigh{5.46} \\
\quad {\scriptsize \(K=8\)} & 5.53 & 5.41 & 5.30 & 5.34 & \errmid{\avgstd{5.40}{0.09}} & 5.46 & 5.39 & 5.52 & 5.43 & \errhigh{\avgstd{5.45}{0.05}} & \errmid{5.39} \\
\quad {\scriptsize \(K=9\)} & 5.42 & 5.30 & 5.40 & 5.38 & \errmid{\avgstd{5.37}{0.05}} & 5.43 & 5.39 & 5.52 & 5.43 & \errhigh{\avgstd{5.44}{0.05}} & \errlow{5.37} \\
\quad {\scriptsize \(K=10\)} & 5.38 & 5.34 & 5.42 & 5.37 & \errmid{\avgstd{5.38}{0.03}} & 5.47 & 5.38 & 5.54 & 5.49 & \errhigh{\avgstd{5.47}{0.06}} & \errlow{5.37} \\
\quad {\scriptsize \(K=11\)} & 5.35 & 5.34 & 5.50 & 5.38 & \errmid{\avgstd{5.39}{0.06}} & 5.47 & 5.36 & 5.53 & 5.43 & \errhigh{\avgstd{5.45}{0.06}} & \errlow{5.36} \\
\quad {\scriptsize \(K=12\)} & 5.36 & 5.35 & 5.56 & 5.37 & \errhigh{\avgstd{5.41}{0.09}} & 5.45 & 5.35 & 5.48 & 5.43 & \errhigh{\avgstd{5.43}{0.05}} & \errlow{5.36} \\
\quad {\scriptsize \(K=13\)} & 5.45 & 5.40 & 5.48 & 5.38 & \errhigh{\avgstd{5.43}{0.04}} & 5.47 & 5.32 & 5.45 & 5.40 & \errhigh{\avgstd{5.41}{0.06}} & \errmid{5.40} \\
\quad {\scriptsize \(K=14\)} & 5.48 & 5.36 & 5.47 & 5.44 & \errhigh{\avgstd{5.44}{0.05}} & 5.49 & 5.36 & 5.46 & 5.43 & \errhigh{\avgstd{5.44}{0.05}} & \errhigh{5.41} \\
\bottomrule
\end{tabular}}
\caption{Mean relative error (MRE) (\%) of category-level mixture estimation under uniform targets. The source-mixture columns follow the same layout as \autoref{tab:token_level_estimation}. DMI is source-independent and therefore shown as a single merged value. Darker green indicates higher error in the QGDE \(K\)-ablation rows.}
\label{tab:appendix_category_level_uniform}
\vspace{-1em}
\end{table*}

\autoref{tab:appendix_category_level_uniform} follows the same layout as \autoref{tab:token_level_estimation}: single-source columns use one known corpus, 70\%-mixed columns use one dominant known corpus with the other three mixed in equally, and the Target-like column uses the same uniform mixture as the target. We report category-level mean relative error, scaled by 100.

\paragraph{QGDE remains below DMI.}
Across both language and domain targets, QGDE gives substantially lower category-level error than the source-independent DMI baseline. In the language setting, DMI has error 9.09, while the QGDE averages are around 3.1--3.3 across source mixtures. In the domain setting, DMI has error 15.14, while QGDE stays around 5.3--5.5.

\paragraph{Anchor gains are weaker after aggregation.}
The \(K\)-sweep is less monotonic than in token-level estimation. For languages, increasing \(K\) improves the early rows but the results quickly cluster near 3.1. For domains, most QGDE rows remain in a narrow band around 5.3--5.5. This supports the observation in \autoref{sec:aggregating_token_ratios} that category-level aggregation smooths and redistributes token-level improvements, so additional anchors have a weaker visible effect after aggregation.

\paragraph{Language mixtures remain easier.}
The language block consistently has lower error than the domain block. This matches the main results: language categories provide more separable token-level evidence, whereas domain categories share more vocabulary and therefore make aggregate mixture estimation harder.

\section{Target-Independent Selection of \texorpdfstring{\(K\)}{K}}
\label{app:heldout-k}

We additionally select \(K\) without observing target-corpus ratios. For each known source, we use three folds: two folds fit the QAC trends and select anchors, and the remaining fold measures held-out error for each \(K\in\{3,\ldots,14\}\). The lowest-error \(K\) is then fixed before target evaluation.

\begin{table}[t]
\centering
\scriptsize
\setlength{\tabcolsep}{3pt}
\renewcommand{\arraystretch}{1.05}
\begin{tabular}{@{}lrrrr@{}}
\toprule
Selection rule & Overall & Language & Domain & SmolLM \\
\midrule
Fixed \(K=14\) & 4.676\% & 3.857\% & 5.087\% & 6.037\% \\
Held-out selected \(K\) & 4.413\% & 3.231\% & 5.162\% & 5.858\% \\
Difference & $-0.263$ pp & $-0.626$ pp & $+0.076$ pp & $-0.179$ pp \\
\bottomrule
\end{tabular}
\caption{Error after target-independent \(K\) selection.}
\label{tab:heldout-k-selection}
\end{table}

As shown in \autoref{tab:heldout-k-selection}, held-out selection is slightly better overall and remains comparable to fixed \(K=14\); the domain average is only 0.076 percentage points higher.

\section{Proxy Corpus Replacement}
\label{app:proxy-replacement}

We keep OSCAR as the target and replace the mC4 proxy used in \autoref{tab:token_level_estimation} with MADLAD-400.

\autoref{tab:proxy-corpus-replacement} shows that replacing mC4 with MADLAD-400 does not weaken QGDE: error decreases in all three source-setting groups. The SmolLM experiment likewise already uses RedPajama-C4, a Wikipedia--arXiv mixture, and CodeParrot as proxies rather than the exact SmolLM components.

\begin{table}[t]
\centering
\footnotesize
\setlength{\tabcolsep}{4pt}
\renewcommand{\arraystretch}{1.05}
\begin{tabular}{@{}lrrr@{}}
\toprule
Proxy corpus & Single-source & 70\%-mixed & Target-like \\
\midrule
mC4 & 5.80\% & 4.78\% & 4.68\% \\
MADLAD-400 & 4.17\% & 3.72\% & 4.02\% \\
Difference & $-1.63$ pp & $-1.06$ pp & $-0.66$ pp \\
\bottomrule
\end{tabular}
\caption{Proxy corpus replacement with OSCAR fixed as the target.}
\label{tab:proxy-corpus-replacement}
\vspace{-1em}
\end{table}

\section{Additional Validation on Pythia}
\label{app:pythia-validation}

We use the GPT-NeoX tokenizer released with Pythia \citep{biderman2023pythia} and compute reference token ratios from a streaming sample of the deduplicated Pile \citep{gao2020pile}. We evaluate QGDE at the target-independent setting $K=14$ with two mixed-domain proxies: a uniform FineWeb--Wikipedia--OpenWebMath--CodeParrot mixture and a pretrain-like mixture.

As shown in \autoref{tab:pythia-validation}, QGDE obtains 4.91--4.93\% error under the two mixed-domain proxies, compared with 10.42--10.64\% for direct transfer and 7.97--8.75\% for PoCTrace. This sampled validation supports the token-level result on a second released BPE tokenizer; it does not include a target-like proxy or category-level evaluation.

\begin{table}[t]
\centering
\footnotesize
\setlength{\tabcolsep}{5pt}
\renewcommand{\arraystretch}{1.05}
\begin{tabular}{@{}lrrr@{}}
\toprule
Source proxy & Transfer & PoCTrace & QGDE \\
\midrule
Uniform & 10.64 & 7.97 & \textbf{4.93} \\
Pretrain-like & 10.42 & 8.75 & \textbf{4.91} \\
\bottomrule
\end{tabular}
\caption{Token-level validation on Pythia. Values are mean relative log-ratio error (\%); lower is better. QGDE uses $K=14$.}
\label{tab:pythia-validation}
\end{table}

\section{Tokenizer-Family Scope}
\label{app:tokenizer-family}

Because QGDE relies on BPE merge-order statistics, we test whether the ID--ratio distribution transfers from mC4 BPE tokenizers to OSCAR tokenizers built with BPE, Unigram, or WordPiece \citep{kudo-2018-subword,song-etal-2021-fast}.

\autoref{tab:tokenizer-family-transfer} shows that transferability is strongest within BPE, weaker for Unigram, and substantially weaker for WordPiece. QGDE is therefore scoped to BPE in this work.

\begin{table}[H]
\centering
\footnotesize
\setlength{\tabcolsep}{5pt}
\begin{tabular}{@{}lrrrr@{}}
\toprule
Transfer setting & En & Fr & Zh & Ja \\
\midrule
BPE $\rightarrow$ BPE & 0.8778 & 0.9245 & 0.9383 & 0.9579 \\
BPE $\rightarrow$ Unigram & 0.7620 & 0.8574 & 0.8461 & 0.8503 \\
BPE $\rightarrow$ WordPiece & 0.6421 & 0.6376 & 0.5018 & 0.6167 \\
\bottomrule
\end{tabular}
\caption{Transferability of token ID--ratio distributions from mC4 to OSCAR.}
\label{tab:tokenizer-family-transfer}
\end{table}

\end{CJK}
\end{document}